\documentclass[a4paper, 11 pt, conference]{ieeeconf}  
\IEEEoverridecommandlockouts                            
\usepackage[a4paper, left=1.59cm, right=1.59cm, top=1.91cm, bottom=2.54cm]{geometry}
\usepackage[english]{babel}
\usepackage{authblk}
\usepackage{amsmath}
\usepackage{graphicx}
\usepackage[colorlinks=true, allcolors=blue]{hyperref}
\usepackage{fancyhdr}
\usepackage{subfiles}
\usepackage{tikz}

\newcommand\copyrighttext{%
  \footnotesize This work has been submitted to the IEEE for possible publication. Copyright may be transferred without notice, after which this version may no longer be accessible. 2023 IEEE. Personal use of this material is permitted. Permission from IEEE must be obtained for all other uses, in any current or future media, including reprinting/republishing this material for advertising or promotional purposes, creating new collective works, for resale or redistribution to servers or lists, or reuse of any copyrighted component of this work in other works.}
\newcommand\copyrightnotice{%
\begin{tikzpicture}[remember picture,overlay]
\node[anchor=south,yshift=10pt] at (current page.south) {\fbox{\parbox{\dimexpr\textwidth-\fboxsep-\fboxrule\relax}{\copyrighttext}}};
\end{tikzpicture}%
}

\fancypagestyle{copyright}{
  \fancyhf{} 
  \fboxsep=10pt 
  \fboxrule=2pt 
  \fbox{\parbox{\dimexpr\textwidth-2\fboxsep-2\fboxrule\relax}{\small\lipsum[1]}} 
}

\begin{document}

\copyrightnotice

\title{\LARGE \bf Object Semantics Give Us the Depth We Need: \\Multi-task Approach to Aerial Depth Completion
{\footnotesize}

}
\author{Sara Hatami Gazani$^{1}$, Fardad Dadboud$^{2}$, Miodrag Bolic$^{2}$, Iraj Mantegh$^{3}$, Homayoun Najjaran$^{1*}$
\thanks{The authors would like to acknowledge the funding from the National Research Council (NRC) Canada under the grant agreement DHGA AI4L-129-2 (CDB \#6835).}
\thanks{*Corresponding author}
\thanks{$^{1}$Department of Mechanical Engineering, University of Victoria, 3800 Finnerty Road, Victoria, BC, Canada, V8P 5C2
        {\tt\small\{sarahatami, najjaran\}@uvic.ca}}%
\thanks{$^{2}$School of Electrical Engineering and Computer Science, University of Ottawa, 800 King Edward, Ottawa, ON, Canada, K1N 6N5
        {\tt\small \{fardad.dadboud, miodrag.bolic\}@uottawa.ca}}%
\thanks{$^{2}$National Research Council Canada (NRC), Montreal, QC, Canada
        {\tt\small iraj.mantegh@cnrc-nrc.gc.ca}}%
}

\maketitle
\thispagestyle{empty}
\pagestyle{empty}

\begin{abstract}
Depth completion and object detection are two crucial tasks often used for aerial 3D mapping, path planning, and collision avoidance of Uncrewed Aerial Vehicles (UAVs). Common solutions include using measurements from a LiDAR sensor; however, the generated point cloud is often sparse and irregular and limits the system's capabilities in 3D rendering and safety-critical decision-making. To mitigate this challenge, information from other sensors on the UAV (viz., a camera used for object detection) is utilized to help the depth completion process generate denser 3D models. Performing both aerial depth completion and object detection tasks while fusing the data from the two sensors poses a challenge to resource efficiency. We address this challenge by proposing a novel approach to jointly execute the two tasks in a single pass. The proposed method is based on an encoder-focused multi-task learning model that exposes the two tasks to jointly learned features. We demonstrate how semantic expectations of the objects in the scene learned by the object detection pathway can boost the performance of the depth completion pathway while placing the missing depth values. Experimental results show that the proposed multi-task network outperforms its single-task counterpart, particularly when exposed to defective inputs.


\end{abstract}

\section{Introduction}
3D mapping and understanding of scenes are crucial tasks for Uncrewed Aerial Vehicles (UAVs) that are responsible for large-scale inspections and condition assessment. While visually inspecting a target such as infrastructure for monitoring or 3D modelling purposes, it is crucial for the UAV to efficiently deploy its onboard sensors to acquire information from its surroundings to ensure a safe, collision-free travel while satisfying the mission objectives. 

\begin{figure}[htbp]
\centerline{\includegraphics[width=\columnwidth]{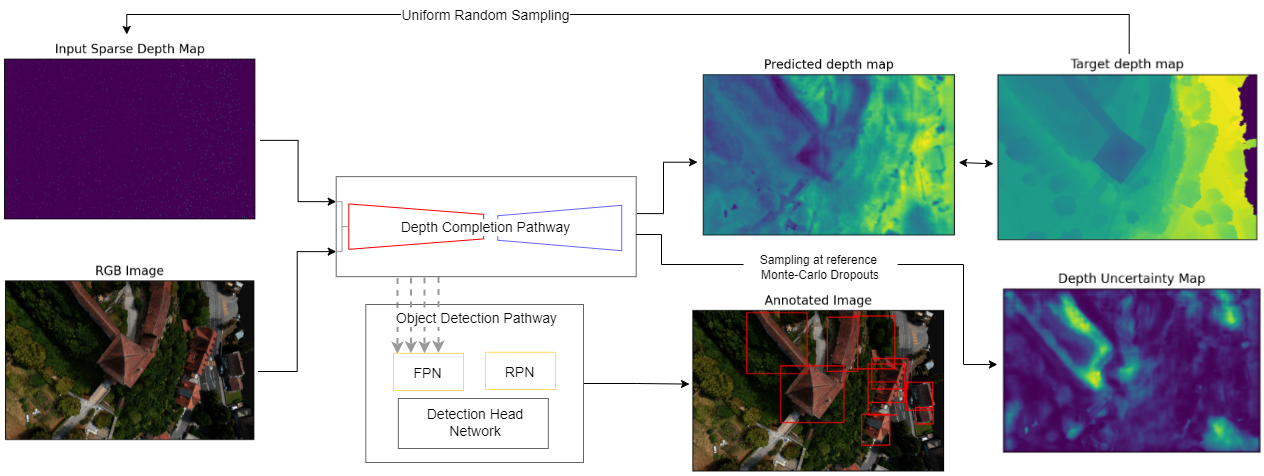}}
\caption{Overview of the proposed multi-task learning framework and the training process. Two tasks of depth completion and object detection are jointly carried out in an encoder-focused multi-task network in a hard parameter sharing fashion to introduce resource efficiency and more robustness to both tasks. }
\label{framework}
\end{figure}

Other than algorithms that are based on Structure-from-Motion (SfM)~\cite{sfm}, using reliable LiDAR sensors that produce high-resolution 3D point clouds is an option; however, their measurements are typically prone to noise, thereby diminishing the confidence of decision-making algorithms. On the other hand, LiDAR sensors with higher performance guarantees may surpass the drones' payload capabilities and bring about power consumption problems. Accordingly, numerous techniques have been devised to address the challenges caused by defects and output sparsity of the sensors that the drones could carry. Among these techniques, depth completion methods tackle this problem by processing sparse depth maps to estimate the missing depth values and output denser depth maps. These methods either use only the depth map (unguided depth completion), or benefit from auxiliary information provided by the RGB images acquired through a camera and a registered depth sensor (RGB-guided depth completion). 

Depth inpainting is one of the early attempts to solve the depth comlpetion problem~\cite{inpainting}. Although similar methods have proven useful for certain purposes, they do not suffice the need for necessary levels of generalizability and scalability required for outdoor inspections. Consequently, owing to their impressive performance, deep learning-based methods have become the main point of interest in the field. The state-of-the-art works present developments in using conventional and modified versions of Convolutional Neural Networks (CNNs) for the task of depth completion~\cite{uncertaintyaware, revisitedcnn}. In both traditional and deep learning-based RGB image-guided methods, the core idea is to benefit from the changes in the local geometric features, texture, and color of the scene to place the missing depth values and even correct the ones already existing in the map. The basic difference in these methods is the type of operations carried out to extract the scene features that help the regression process. Respectively, the choice of the right geometric features to extract from the RGB image to guide the depth completion process in a deep learning framework remains an open research challenge. We address this challenge and reflect on the typical objectives of a UAV inspection flight by proposing a multi-task network that achieves both depth completion and object detection in a single pass. We explore how incorporating object detection as an auxiliary task impacts the results of the depth completion as opposed to accomplishing the task using a single-task network. Object detection is chosen to represent a widely used application of UAV inspection flights and an inexpensive alternative to pixel-wise annotation and processing required for semantic segmentation, which is usually impractical for aerial data acquisition. Additionally, to study the change in the behavior of the network with employing the object detection pathway, we use the Monte-Carlo dropout technique~\cite{mcdrop} to generate pixel-wise uncertainty maps for both cases.

\section{Motivation} \label{motivation}
Most of the work in traditional and deep learning-based depth completion focuses on scene representations containing features that are local to the individual pixels, ignoring the neighborhood of the pixel which, along with the pixel, make an entity of a semantic object. In such cases, texture and color information would be utilized to account for the presence of the same object along a surface~\cite{inpainting}, or the network would work to extract specific geometric features such as object boundaries and edges~\cite{boundary} or surface normals~\cite{normal} which requires added computations and supervision. This, in turn, hints that the extracted features would always lack a relevance to the certain spatial relationships existing in similar entities of semantic objects belonging to the same category. 
In a more practical context considering a UAV inspection flight, if the network identifies an object, e.g. a bridge, the network can use its contextual knowledge of the bridge's probable shape and size to estimate the expected depth values for pixels along the sides of it. This will also pave the way for the network to reach a more complete 3D reconstruction of the scene in case occlusion poses a challenge. With this as the core idea of the presented study, we propose to guide the extracted features along the depth completion stream to satisfy a second task, namely object detection, through a multi-level shared backbone. Figure~\ref{framework} represents an overview of the model. 

 \begin{figure*}[ht]
\centerline{\includegraphics[scale=0.32]{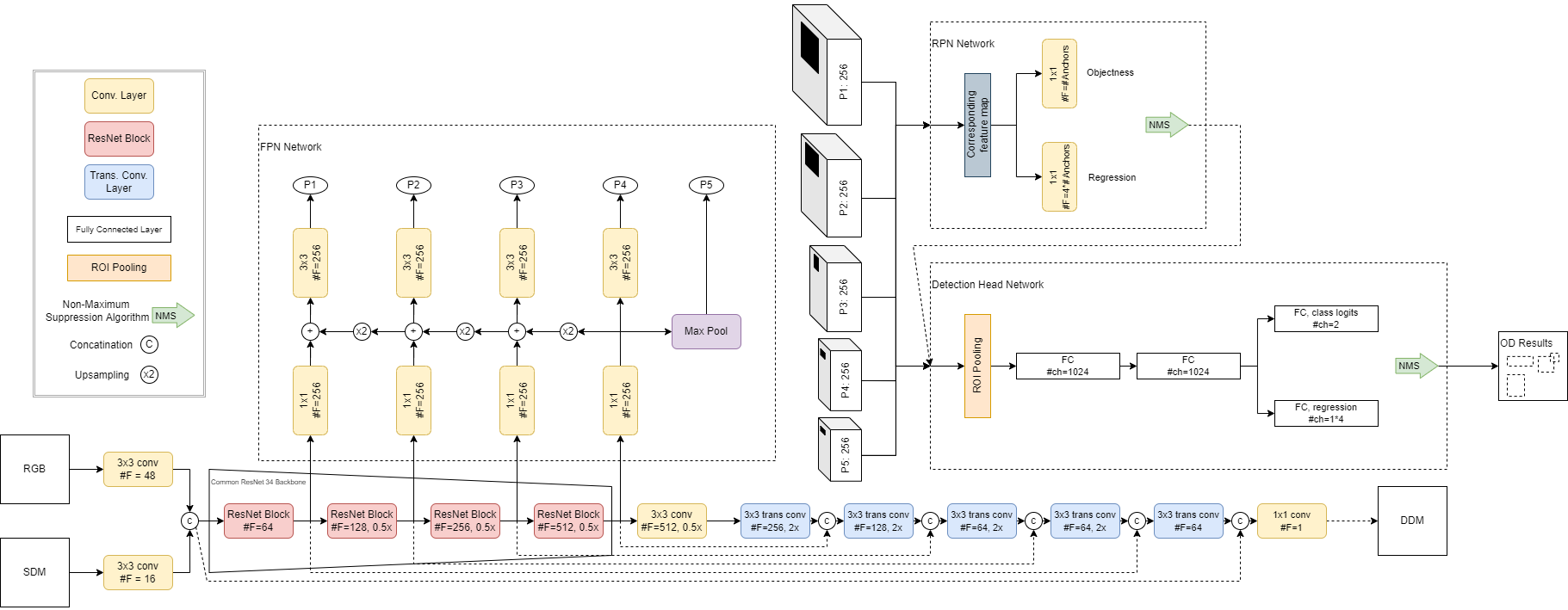}}
\caption{The proposed architecture for the multi-task network. Features from different stages of encoding the fused data from the RGB image and the depth map are shared between the object detection pathway and the depth completion pathway.}
\label{architecture}
\end{figure*}

\section{Related Work}\label{relatedwork}
\subsection{Deep Learning-based Depth Completion}
Previous works that use RGB guidance for depth completion often aim to develop a network that learns explicit geometric features of the scene to help improve the quality of the completed depth map. These features can be either 2D or 3D representations of the scene which capture specific characteristics of the surface that help the network interpolate and predict more reliable depth values. Among such representations are surface normals that have received reasonable attention as they propose an added geometric constraint for the depth completion network~\cite{normal, deeplidar, indoor}. \cite{deepdepth} explored different depth representations that contribute to depth completion and found the normals along surface boundaries to be most effective. In \cite{Ma2017SparseToDense}, a novel encoder-decoder-like model based on the well-known ResNet architecture \cite{resnet} was deployed to directly derive a dense depth map from the input RGB image and sparse depth map. Due to the efficiency and comprehensibility of this architecture, we found it most suitable for integrating with a second stream, i.e. the  object detection stream, for the purposes of this study.

\subsection{Multi-task Networks for Depth Completion}
Multi-task learning is a promising machine learning area which takes advantage of auxiliary tasks for resource efficiency and reducing computation cost at inference time. Depending on the relevancy of the target tasks handled by a single network or different networks merged together, the performance quality of the network can experience an improvement or a decline compared to the case of using separate single-task networks. In this regard,~\cite{standley2020tasks} studied the efficacy of grouping different sets of tasks in computer vision to solve through multi-task learning. By guiding the network's loss function to suit a main task and an auxiliary task, they demonstrated improved results for the main task, which could come at the cost of degraded results for the auxiliary task. Although the use of multi-task networks prevents re-learning certain features for closely-related tasks and proves its efficiency in the use of resources, it comes with the trade-off between the model performance and resource gain~\cite{bailer2019resource}. While choosing the tasks to be included in the network, it is also necessary to consider the requirements for the corresponding application. In recent years, multi-task learning together with multi-sensor fusion have been used in 2D and 3D perception systems to improve the computation cost at inference time and the system performance. In~\cite{liang2019multi}, a joint model was proposed which successfully combined four tasks, with 3D object detection as the target task to be improved, which demonstrated the complementary natures of mapping, object detection, and depth completion. They trained a multi-task multi-sensor detector in an end-to-end fashion which achieved dense point-wise and ROI-wise feature fusion. \cite{zou2020simultaneous} improved both tasks of semantic segmentation and depth completion by using a feature-sharing encoder and a three-branch decoder.

\section{Methodology}
The work is executed by employing separate sub-networks, referred to as pathways or streams, in a multi-task model where extracted features from several stages of encoding throughout the backbone are fed to two task-specific heads. 
While the depth completion network tends to extract features more local to the pixels and of geometric value, the object detection stream extracts features that are more unique to the objects in the scene with higher semantic value. Thus, the parameter sharing allows the backbone to learn a representation space that benefits both tasks, owing to the added supervision. As well, in order to further study the behavior of the multi-task network compared to its single-task opponent, we need a representation for the uncertainty level of the depth maps generated by each of the networks. This is obtained by using Monte-Carlo dropouts inspired by~\cite{mcdrop}. The following sections elaborate on the separate pathways, the feature sharing method for multi-task learning, and the corresponding pixel-wise uncertainty estimation. The details of the model architecture are shown in Fig.~\ref{architecture}. 
\subsection{Depth Completion Pathway}
The core architecture for the depth completion pathway is inspired by~\cite{Ma2017SparseToDense} and follows an encoder-decoder structure. First, the sparse depth map and the RGB image are fed to two different convolutional layers with filters of 16 and 48 channels. The resultant feature maps are concatenated and the merged features go through five different stages of down-sampling using residual blocks of ResNet-34~\cite{resnet} and a convolutional layer with 512 channels. The encoded features are then up-sampled using five transpose-convolutional layers of 256, 128, 64, 64, and 64 channels, respectively. The outputs of each of the down-sampling stages from the encoder are concatenated with the resultant features of the corresponding up-sampling stages in the decoder and are fed to the next stage. The output is a dense depth map image of the same size as the inputs.

\subsection{Object Detection Pathway}
The object detection pathway is designed to serve as the auxiliary stream in this multi-task learning scheme. Inspired by \cite{lin2017feature}, the pathway uses a ResNet-based Faster R-CNN detector augmented by a Feature Pyramid Network (FPN) neck. Intermediate feature maps from the shared backbone are fed to the pathway on 4 different levels. The FPN pulls out the ResNet blocks' outputs and provides 5 different scale feature maps for the next modules. The FPN features are then fed into a Region Proposal Network (RPN) where objectness probability and box regression proposals are produced. These feature maps from the FPN are then used to predict bounding boxes corresponding to the different scale they are generated with. The different anchor boxes from corresponding feature maps are fed to the RPN to provide the proposals. The final detection head then refines the proposed bounding boxes through a Region of Interest (ROI) pooling layer. The outputs of the RPN and the final detection head are then fed to a Non-maximum Suppression (NMS) algorithm to remove the duplicate bounding boxes.

\subsection{Model Uncertainty Representation}\label{uncertainty_sec}
In order for a comparative study of the behaviors of the single- and multi-task networks, we aim to quantify the uncertainty in the models outputs inspired by the work in~\cite{dropout}  where a Bayesian approximation for model uncertainty was obtained using active dropouts at inference time. We use the same method to obtain the predictive mean and variance for the model output by fitting a Gaussian distribution to the results of multiple forward passes. Subsequently, for each pixel in the map, we visualize a corresponding variance value in the uncertainty map, as shown in Fig.~\ref{uncertainty}, as well as a mean to represent the inferred depth result. 
\begin{figure}[tbp]
\centerline{\includegraphics[width=\columnwidth]{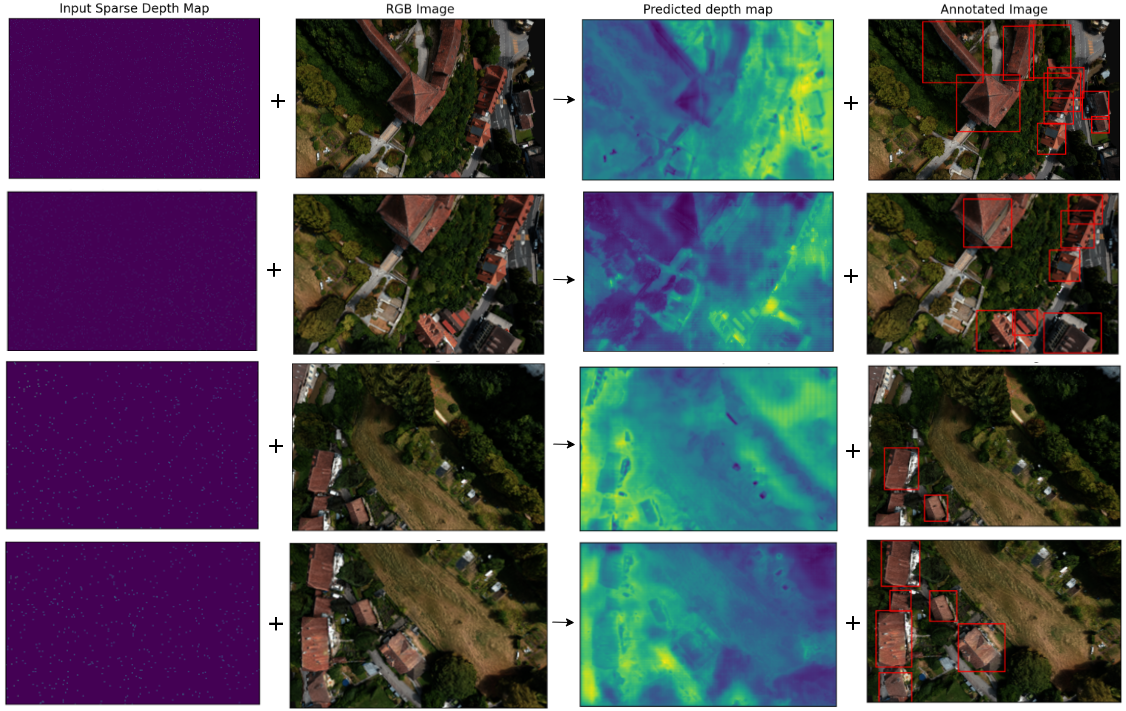}}
\caption{Depth completion and object detection results for the proposed multi-task network.}
\label{result}
\end{figure}

\subsection{Learning Objectives}
\subsubsection{Depth Completion Pathway Learning Objectives}
The learning objectives for this network target the quality of the generated dense depth map in different aspects, for each of which a loss function is defined. 
\\\textbf{Depth Consistency:}
A dense depth map provides direct supervision for the network by acting as the ground truth. In this work, L2-norm, or the Root Mean Squared Error (RMSE), is used to keep track of the consistency of the completed depth given by Eq. \ref{l2}:
\begin{equation}\label{l2}
    l_{consistency} = \frac{1}{n}\sum_{i=1}^{n}{||\hat{Y_i}-Y_i||_2}
\end{equation}
where n is the number of pixels and $Y_i$ and $\hat{Y_i}$ represent predicted and ground truth depth values of the $i^{th}$ pixel, respectively.
\\\textbf{Depth Smoothness:}
To guarantee better depth completion quality, it is important to reduce the noise in the depth map by taking into account the gradient of the depth values in the x- and y-axis on the plane of the depth map. That is implemented by taking the L1-norm of the second order derivative of the depth values. This ensures that any irregularities in the depth map are suppressed (Eq. \ref{smoothness}). 
\begin{equation}\label{smoothness}
    l_{smoothness} = \frac{1}{n}\sum_{i=1}^{n}{| \partial_x^2 \hat{Y_i}| + |\partial_y^2 \hat{Y_i}|}
\end{equation}

\subsubsection{Object Detection Pathway Learning Objectives}
Based on the original Faster R-CNN loss \cite{ren2015faster} and its implementation in TorchVision~\cite{TorchVision} detection framework, we utilized a detection loss which contains two terms: proposal loss and final detection loss which are summed with a relative weight $\lambda$ (Eq. \ref{ODLoss}).
\begin{equation}\label{ODLoss}
    l_{detection} = l_{proposal} + \lambda l_{final~detection}
\end{equation}
We refer the reader to \cite{ren2015faster, pytorch} for more details on the loss function of this pathway.
The proposed network is trained with a weighted combination of the losses for the two streams.

\begin{figure}[tbp]
\centerline{\includegraphics[width=\linewidth]{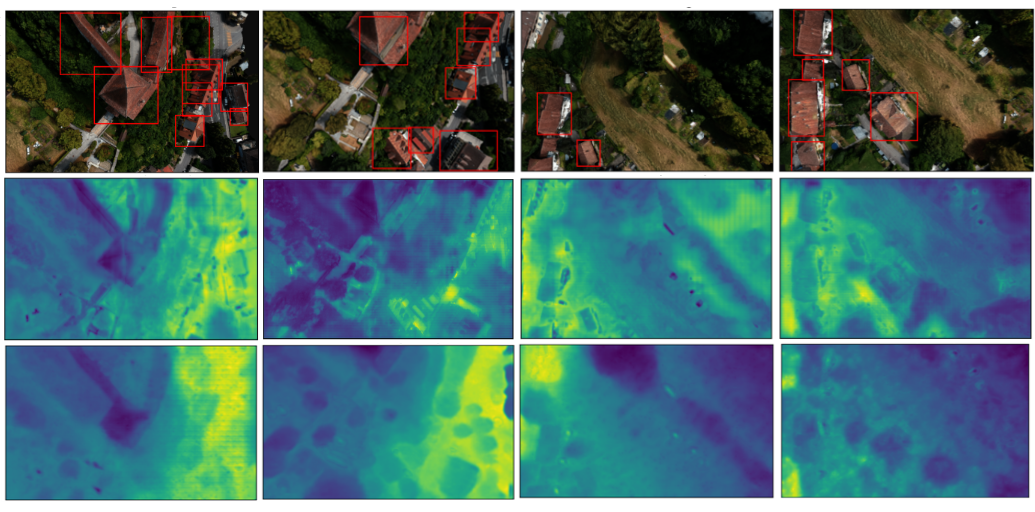}}
\caption{Comparison between the output depth maps of the single- and the proposed multi-task networks. Notice the difference in the object boundaries. The multi-task network (second row) identifies comparatively crisp boundaries in areas where an object has been detected. The single-task network (third row), in contrast, predicts blurred depths.}
\label{comp}
\end{figure}

\section{Experiments and Results}\label{SCM}
\subsection{Implementation Details}
The proposed network including its two pathways is implemented in PyTorch~\cite{pytorch} using the Lightning modules~\cite{Falcon_PyTorch_Lightning_2019}. We used Adam as the optimizer with a starting learning rate of $10^{-4}$ which is multiplied by 0.5 after every 5 epochs. We used a NVIDIA GTX 3060 with up to 16 GB of memory to train the networks for 20 epochs. The input images (RGB and depth map) were fed to the network after being resized to $320\times240$ in separate batches of size 1. The sparse depth maps were generated by uniformly sampling $0.7\%$ of points from the dense ground truth depth maps. 

\subsection{Dataset}
We use the Aerial Depth Dataset developed by~\cite{aerial} which includes RGB and depth images of infrastructure models generated using a photogrammetry software. As no annotations were available to perform the object detection task on this dataset, we manually annotated the buildings and bridges in 1000 frames. We also chose not to display the labels associated with the detected objects in the results since our focus is rather on the locations of the objects in the scene than on their specific categories.

\begin{figure}[tbp]
\centerline{\includegraphics[width=\linewidth]{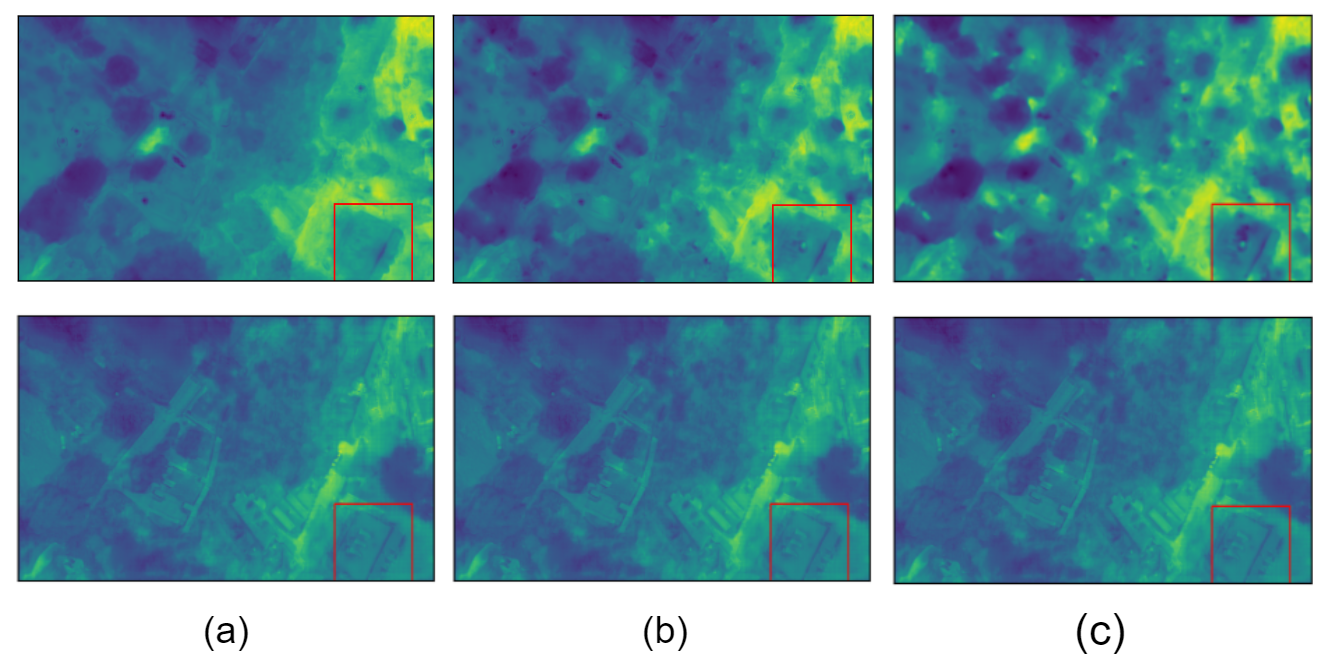}}
\caption{Comparing the noise robustness of the single- and multi-task networks. Gaussian distance-dependent noise is applied to the input sparse depth maps with increasing variance from (a) $10\%$ to (b) $20\%$ and (c) $40\%$ of the actual values. First and second row show the results of the single-task and the multi-task networks, respectively.}
\label{noise}
\vspace{-2.5em}
\end{figure}

\subsection{Comparative Analysis}
To test our hypothesis mentioned in Section~\ref{motivation}, we trained and tested the single-task and the proposed multi-task networks with the same hyper-parameters to compare the quality of the output depth map and their robustness in response to defective inputs. Figure~\ref{result} demonstrates the input-output summary of the multi-task network and the results are compared to those of the single-task network as depicted in Fig.~\ref{comp}. From a qualitative perspective, it can be seen that the output depth map from the multi-task network has done a better job at replacing depth values around the objects in the scenes that have been annotated in the ground truth images, which is apparent from clear object boundaries in the generated depth maps.

\subsection{Robustness Analysis}
\begin{figure}[tbp]
\centerline{\includegraphics[width=\linewidth]{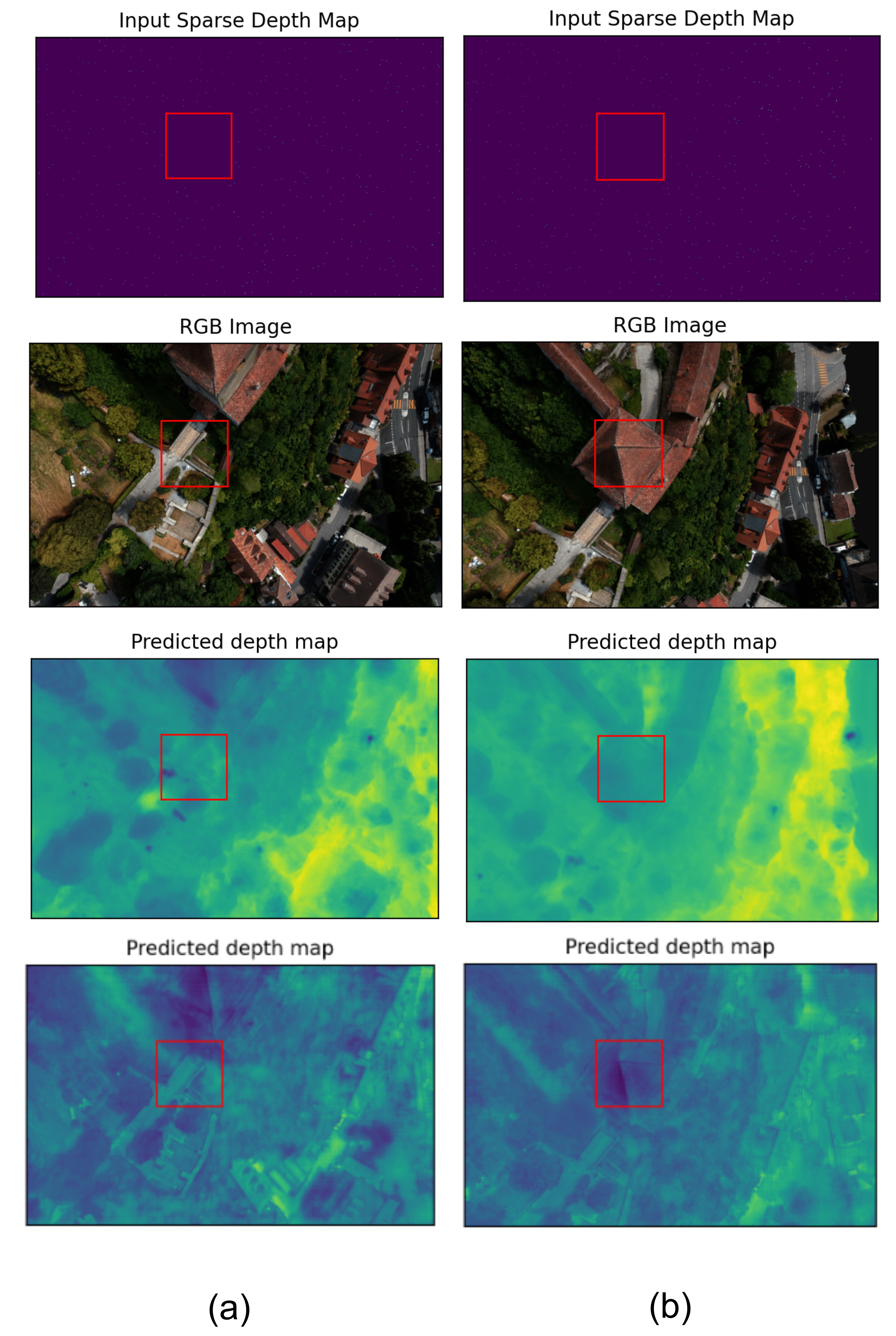}}
\caption{Comparing the performance of the two networks in response to incomplete sparse maps as inputs with missing values in the marked box for two different (a) and (b) scenes. The third row corresponds to the outputs of the single-task network and the fourth row demonstrates the outputs of the multi-task network.}
\label{robust} 
\vspace{-1.5em}
\end{figure}

As discussed before, the added supervision guides the shared backbone layers to extract contextual information from the inputs, ultimately leading to enhanced performance in depth completion. To investigate how this context could improve the robustness of the proposed network, we expose the single- and multi-task networks to defective input depth maps. First, we subject the networks to RGB images with sparse depth maps that have increased Gaussian distance-dependent noise and demonstrate the results in Fig.~\ref{noise}. Then, random boxes are introduced on the maps which result in missing depth values within the boxes to test if the relative depth of the structures can be inferred even in the absence of their actual depth values. Figure~\ref{robust} shows two samples of the results of this experiment. From a qualitative standpoint, it is evident that the multi-task network is more effective at predicting missing depth values, particularly those associated with structures, compared to the other network.

\subsection{Uncertainty Map Analysis}
\begin{figure}
\centerline{\includegraphics[width=\linewidth]{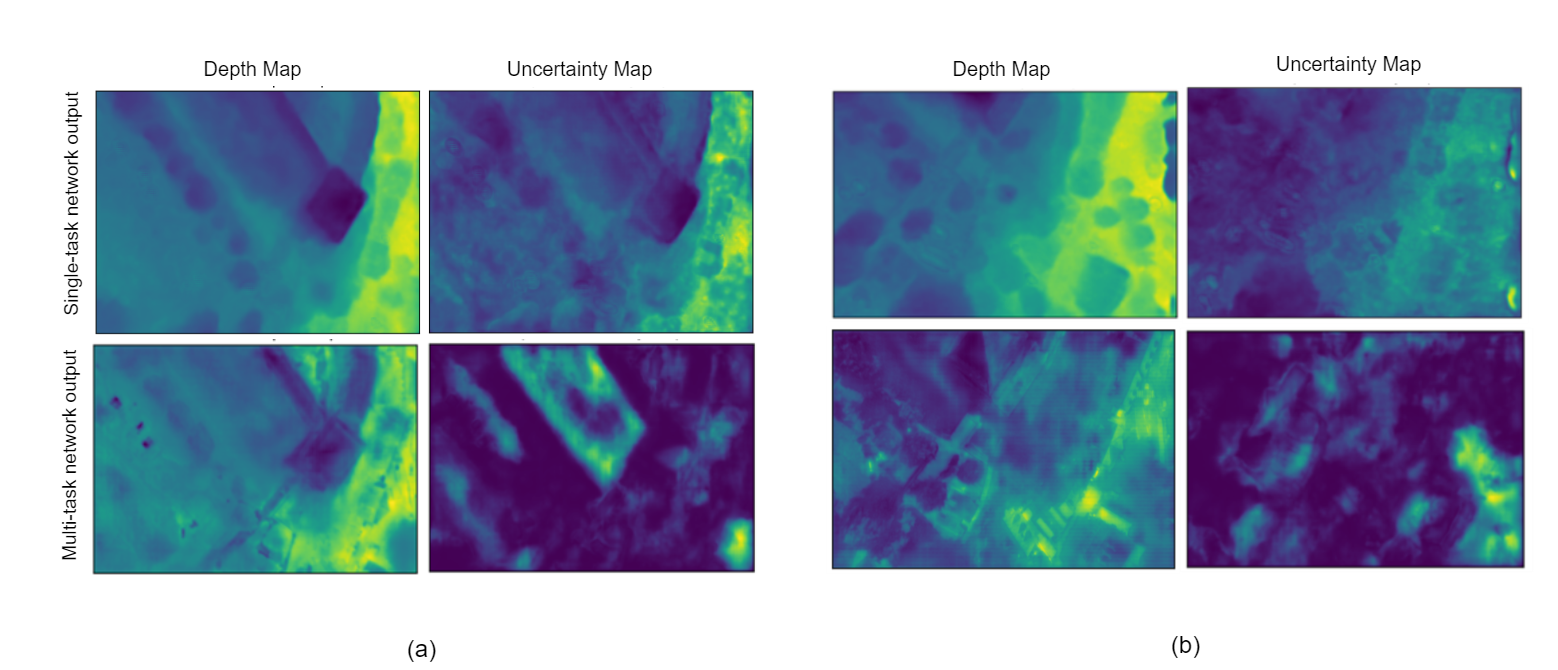}}
\caption{Uncertainty map comparison between the depth maps generated by the single- and proposed multi-task network for two different (a) and (b) scenes. The first columns are the output depth maps and the second columns are the uncertainty maps for the singe- and multi- task network results in the first and second rows.}
\label{uncertainty}
\vspace{-1.5em}
\end{figure}

Figure~\ref{uncertainty} demonstrates the low variance in uncertainty outputs of the multi-task network when compared to the single-task counterpart. On visual inspection, the uncertainty maps of areas predicted between the bounding boxes, e.g., buildings, present significantly less uncertainty. The uncertainty estimation through the proposed network for unannotated areas e.g., trees, remains higher than the annotated areas. In contrast, with the single-task network, objects like buildings have higher uncertainty which is detrimental to UAV inspection missions.

\section{Conclusion and Future Work}
To address the issue of sensor data defects that can impact both the 3D modelling and safety insurance of autonomous aerial vehicles, we proposed and implemented a multi-task network to attain the tasks of depth completion and object detection. We introduced semantic feature maps as representations of the scene based on which the depth completion results can be improved. The network runs on a shared backbone with two task-specific heads assigned to respectively produce a dense depth map and bounding boxes localizing the infrastructure present in the scene. A comparative analysis was carried out on the results of the single- and the proposed multi-task networks which proved enhanced performance of the multi-task network in producing depth maps for the pixels belonging to the detected objects by the object detection stream. While this work primarily serves as proof of concept, our future research will focus more on the inspection and explanation of the shared feature maps and extending the annotated dataset to enable higher generalizability to unseen infrastructure. It is also worth mentioning that the results of this network can be used for online view planning problems where detection of the target as well as its relative position to the UAV in the form of a dense depth map is essential.  

\bibliographystyle{IEEEtran}
\bibliography{main}

\end{document}